\documentclass{pprai}

\usepackage[utf8]{inputenc}
\usepackage[T1]{fontenc}
\usepackage{graphicx}
\usepackage{amsthm}
\usepackage{txfonts}
\usepackage{url}
\usepackage{algpseudocode}
\usepackage{subcaption}

\usepackage[final]{changes}

\title{Mitigating Dimensionality in 2D Rectangle Packing Problem under Reinforcement Learning Schema}
\headtitle{Mitigating Dimensionality in 2D Rectangle Packing Problem...}

\author{Waldemar Ko{\l}odziejczyk$^{1[0000-0002-2031-4264]}$, Mariusz Kaleta$^{2[0000-0002-2225-8956]}$}
\headauthor{W. Ko{\l}odziejczyk, M. Kaleta}
\affiliation{%
  $^1$
  kolodziejczykwaldemar222@gmail.com\\
  $^2$Warsaw University of Technology\\
  Faculty of Electronics and Information Technology\\
  ul. Nowowiejska 15/19, 00-665 Warsaw, Poland\\
  mariusz.kaleta@pw.edu.pl}


\begin{document}
\maketitle

\begin{abstract}
This paper explores the application of Reinforcement Learning (RL) to the two-dimensional rectangular strip packing problem. We propose a reduced representation of the state and action spaces that allow us for high granularity. Leveraging UNet architecture and Proximal Policy Optimization (PPO), we achieved a model that is comparable to the MaxRects heuristic. However, our approach has great potential to be generalized to non-rectangular packing problems and complex constraints.

\end{abstract}

\section{Introduction}
We consider a classic NP-hard problem
, yet very practical in many fields
, the rectangle two-dimensional strip packing problem
.  The set of $N$ rectangles is to be packed as densely as possible in a strip of a given width. For simplicity, without loss of generality, we limit the height of the strip to a sufficiently large value and consider a bin of fixed dimensions throughout the paper. We focus on the online version of the problem, where rectangles are processed in descending order of their area. The typical known approaches to solving the problem include constructive heuristics and metaheuristics, among others \cite{duan2018multitask, faroe2000gls}. We aim to apply the Reinforcement Learning (RL) approach, assuming a grid representation of the bin and modeling the problem as a Markov Decision Process (MDP). The grid representation means that the bin is discretized into cells, $w$ columns and $h$ rows, that resemble pixels, and the whole method can be perceived as vision-based. Our primary motivation is that such an approach, in contrast to known heuristics, has a great potential to be generalized to other shapes and to accommodate various constraints. Since the straightforward approach suffers from the dimensionality of the action space, we propose a representation that highly reduces the size of the space.

\section{Literature Review}
A literature review reveals that the practical application of machine learning, including reinforcement learning, in the domain of Bin Packing and Strip Packing is still in its early stages. Traditional approaches to solving these problems involve heuristics such as genetic algorithms \cite{duan2018multitask} and Guided Local Search algorithms \cite{faroe2000gls}. Recent research is primarily focused on RL applied to 3D Bin Packing Problems; however, results are obtained for relatively low granularity of the problem \cite{Zhao2021-we, Zhao2020-at, Puche2022-xa}. 2D Rectangular Strip Packing Problem significantly differs from 3D Bin Packing in operations, model construction, and application contexts. Directly applying Reinforcement Learning to solve 2D rectangular packing has seen limited research \cite{Fang2023-ay}. Nevertheless, recent attempts leveraging machine learning for 2D rectangular packing have yielded notable progress. One strategy involves employing a Convolutional Neural Network as a Q-value estimator within the framework of Double Deep Q Learning \cite{kundu2019-2D-DRL}. However, the effectiveness of the proposed methodology is limited by the very low state resolution of the 6x6 grid. Xu et al. use RL and pointer network for 2D rectangular strip packing problem to determine a sequence of items that are packed with MaxRects heuristic \cite{xu2020transfer}. A promising direction is the integration of reinforcement learning with mathematical optimization models for packing and the exploration of hybrid RL algorithms \cite{Zhao2020-at, Fang2023-ay}. Frequently used algorithms are often not scale invariant and encounter difficulties in transferring to other similar tasks. To the best of our knowledge, this paper is the first to try to solve a high-granularity problem and tackle the problem of action space size.

\section{Model of Markov Decision Process}

Let the tuple $(\mathcal{S}, \mathcal{A}, \mathcal{P}, \mathcal{R})$ represent the Markov Decision Process, where $\mathcal{S}$ is the set of states that an agent can observe, $\mathcal{A}$ is the set of actions of an agent, $\mathcal{R}$ is the reward function, and $\mathcal{P}$ is probability of transition from given state to another state under given action. A pixel-like representation of the bin means large state and action spaces (any location in the 2D plane) \cite{Zhao2020-at}. Our experiments with such representation revealed unsatisfactory performance and convergence issues. To address this challenge, we narrowed down the degrees of freedom along the height dimension. This reduction compelled the agent to operate in a 1D environment, facilitating item placement solely based on the X-coordinate, akin to the mechanics observed in Tetris.

\begin{figure}[ht]
    \centering
    \begin{subfigure}{.24\textwidth}
        \centering
        \includegraphics[width=0.95\linewidth]{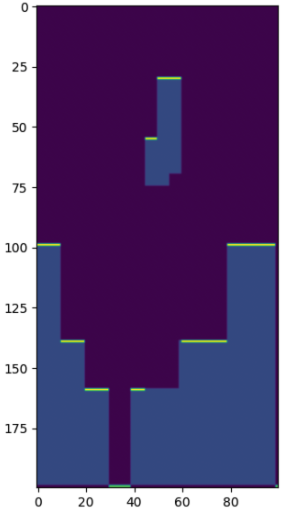}
        \caption{}
        \label{fig:1d_view}
    \end{subfigure} %
    \begin{subfigure}{.24\textwidth}
        \centering
        \includegraphics[width=0.95\linewidth]{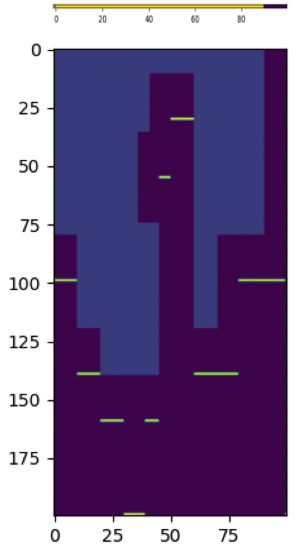}
        \caption{}
        \label{fig:1d_feasibility_map}
    \end{subfigure}%
    \caption{Renders of the exemplary episode after insertion of some elements. (a) Occupied 2D area marked as blue region, channel 1: occupancy vector values represented by heights of yellow bars (b) Available 2D area accessible for new item, channel 2: feasibility binary map drawn above the bin.}
    \label{fig:1d_view_feasibility}
    \end{figure}

Inspired by work \cite{Zhao2020-at} on the 3D Bin Packing Problem, we define the state space $\mathcal{S}$ with five vectors (channels). The reduced state representation encapsulates the spatial configuration of elements in the bin, providing information about the current arrangement, possible placement locations, and details about the shape and dimensions of each element. Channel 1 is a normalized height map $M$ representing the occupation level of the bin. For each pixel, it is the distance from the bottom of the bin to the last encountered item placed in the bin marked as a yellow bar in Fig. \ref{fig:1d_view}. Channels 2-3 are binary masks indicating potential locations, so-called feasibility maps, for placing an element at two different rotation angles: 0 and 90 degrees (see a bar on the top of Fig. \ref{fig:1d_feasibility_map} as exemplary feasibility map for a not rotated object). Channels 4-5 are 2-element embedding representing the shape of the element, including the normalized height and width of the current element. The size of the space is $5\cdot w$.

The action space $\mathcal{A}$ for the RL agent is a tensor composed of two vectors, channel 1 and channel 2. Each channel is a vector of policy network probabilities representing the desirability of placing an element in a specific location; however, channel 1 considers a non-rotated item, and channel 2 considers an element rotated by 90 degrees. The size of the action space is $2\cdot w$.

\begin{figure}[ht]
    \centering
    \includegraphics[width=0.5\textwidth]{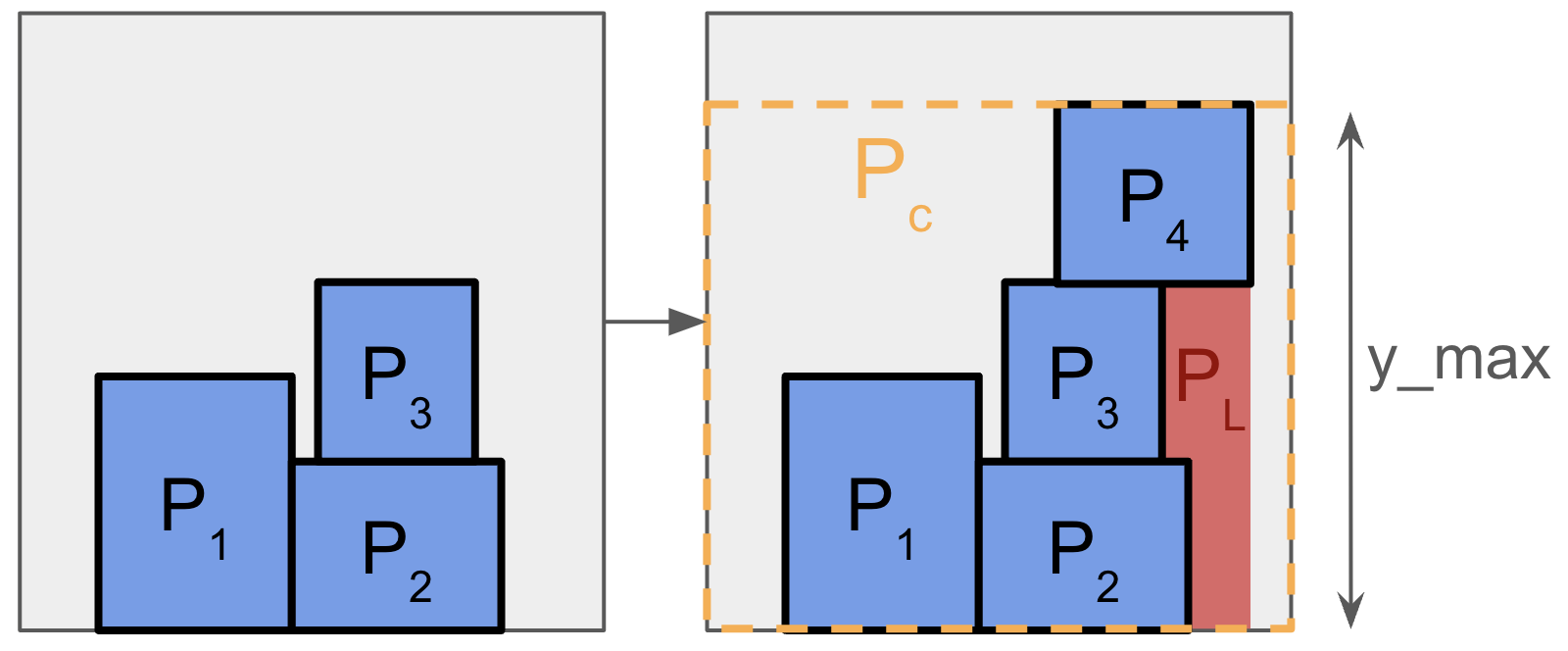}
    \caption{Reward function for 1D environment}
    \label{fig:reward_1d}
    \end{figure}

In order to guide the model towards a desired outcome, we designed two versions of the reward function, (a) $V1$ containing only terminal reward, and (b) $V2$ enhanced with intermediate reward (illustrated in Figure \ref{fig:reward_1d}):
    \begin{equation}
    \texttt{(a)} \ R_T=\frac{\sum_{n=0}^{N-1}P_n}{P_{c}}, \quad \texttt{(b)} \ R_t= \left\{  \begin{array}{ll}
    - P_L, & \textrm{if $t < T$}\\
    \frac{\sum_{n=0}^{N-1}P_n}{P_{c}}, & \textrm{otherwise}
    \end{array} \right.
    \label{eq:reward_1d_v1}
    \end{equation}
where $t$ is iteration, $P_n$ is an area of the $n$-th element, $P_c$ is an area of the region (0, $w$, 0, $y_{max}$) in which elements are present, $P_L$ lost area during step $t$.

\section{Experiment Results}

We tested 500 episodes, each consisting of 15 items, either fixed-sizes or randomly generated, and arranged in descending order by area, following a common practice observed in other BPP heuristics. We assumed $w=125$ and $h=150$. Utilizing Proximal Policy Optimization (PPO), we employed a 1D UNet architecture as the policy network for the RL agent, chosen due to its superior performance and faster convergence compared to the equivalent Deep Q Learning alternative. The selection of the UNet architecture was motivated by the observation that determining optimal probabilities for the best action at any given time can be analogized to a classical Computer Vision segmentation task, with a focus on spatial bias and correlation among neighboring pixels. The stopping criteria for an episode included reaching the end of the item collection or encountering insufficient space for any rotation, applying to both feasibility masks.

\begin{figure}[ht!]
    \centering
    \begin{subfigure}{.24\textwidth}
        \centering
        \includegraphics[width=0.99\linewidth]{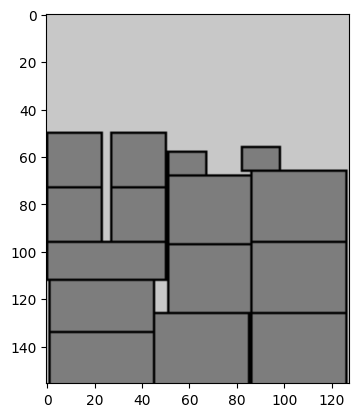}
        \caption{}
        \label{fig:1D_finite_v1}
    \end{subfigure} %
    \begin{subfigure}{.24\textwidth}
        \centering
        \includegraphics[width=0.99\linewidth]{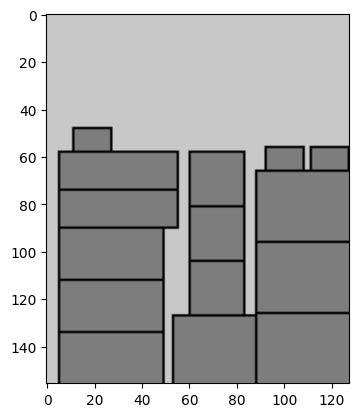}
        \caption{}
        \label{fig:1D_finite_v2}
    \end{subfigure}%
    \begin{subfigure}{.24\textwidth}
        \centering
        \includegraphics[width=0.99\linewidth]{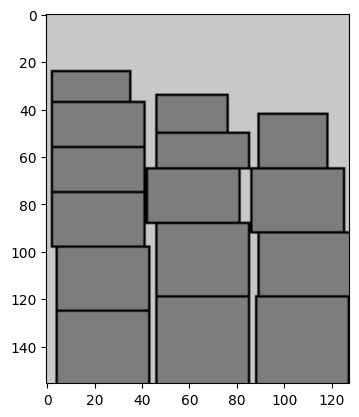}
        \caption{}
        \label{fig:1D_random_v1}
    \end{subfigure}%
    \begin{subfigure}{.24\textwidth}
        \centering
        \includegraphics[width=0.99\linewidth]{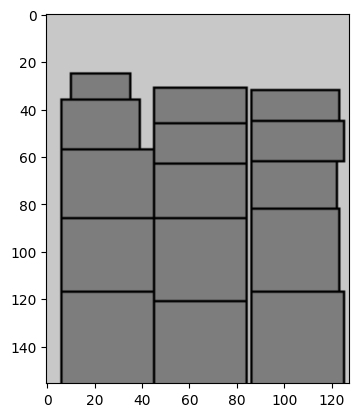}
        \caption{}
        \label{fig:1D_random_v2}
    \end{subfigure}%
    \caption{Exemplary results using (packed elements are white): (a) Finite set of elements with only terminal reward; (b) Finite set of elements with intermediate and terminal reward; (c) Random set of elements with only terminal reward; (d) Random set of elements with intermediate and terminal reward.}
    \label{fig:1D_examples_v1_v2}
    \end{figure}

\begin{figure}[ht!]
    \centering
        \begin{subfigure}{.49\textwidth}
        \centering
        \includegraphics[width=0.9\textwidth]{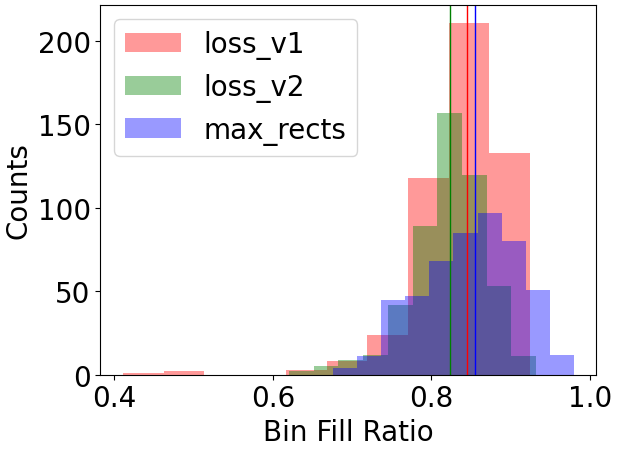}
        \caption{}
        \label{fig:hist_finite}
        \end{subfigure}
        \begin{subfigure}{.49\textwidth}
        \centering
        \includegraphics[width=0.9\textwidth]{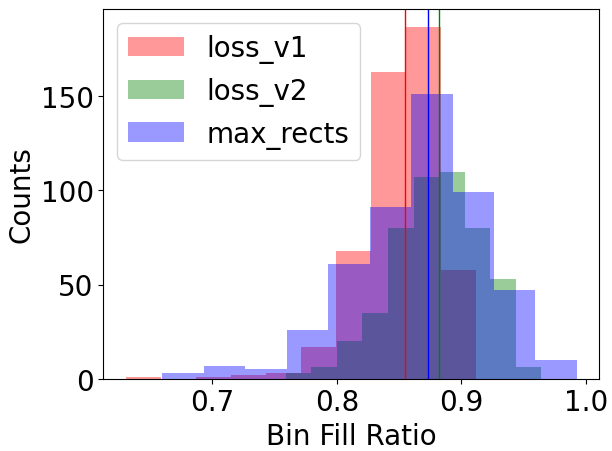}
        \caption{}
        \label{fig:hist_random}
        \end{subfigure}
    \caption{Histograms comparing two reward functions against the MaxRects heuristic for (a) finite set of elements, (b) random set of elements.}
    \label{fig:hist}
    \end{figure}

In the scenario with fixed-size elements, the agent, guided solely by the terminal reward $V1$ (see Fig. \ref{fig:1D_finite_v1}), showcased an ability to plan arrangements along both the right and left borders, minimizing unused space. Figure \ref{fig:1D_finite_v2} highlights a deliberate decision by the agent to leave an unoccupied spot in the middle, strategically mitigating penalties associated with the lost area. However, this strategic choice might lead to a reduced bin-filling ratio by the end of the analyzed episode. The statistical analysis of all examined episodes, when compared to the MaxRects algorithm (see Fig. \ref{fig:hist_finite}), indicates that the terminal reward $V1$ outperformed the intermediate alternative $V2$. The results achieved were slightly inferior to the MaxRects approach.

For the random set of elements scenario, intriguingly, the agent operating with intermediate reward $V2$ achieved superior average results when compared to the alternative utilizing only terminal reward $V1$, outperforming the MaxRects algorithm (Figure \ref{fig:hist_random}). When evaluating distribution properties, it is worth mentioning that both versions, $V1$ and $V2$, exhibit smaller variances compared to the heuristic competitor. This indicates greater stability and reduced uncertainty in the obtained results. Analyzing the results of the exemplary episodes depicted in Figure \ref{fig:1D_random_v1} and Figure \ref{fig:1D_random_v2}, we observe that the agent tends to leave more blank space on the borders. This behavior may be a reflection of uncertainty regarding the next element, a factor that cannot be inferred from experience, as in the fixed-size variant.

\section{Conclusions}

While we find it hard to achieve reasonable models for full 2D bin representation, our reduced space size 1D approach under UNet-based PPO agents resulted in the model comparable to the MaxRects and even outperforming it in some cases. This achievement is promising for further investigation of non-rectangular packing. Further exploration of scenarios and variations in problem settings may provide additional insights into the capabilities and limitations of RL in the context of 2D rectangular strip packing problems. We also believe that a promising avenue lies in combining heuristics with RL-based sorting or integrating multiple heuristics.


\end{document}